\documentclass[10pt,twocolumn,letterpaper]{article}

\usepackage{iccv}
\usepackage{times}
\usepackage{epsfig}
\usepackage{graphicx}
\usepackage{amsmath}
\usepackage{amssymb}
\usepackage{amsthm}
\usepackage{mathtools}
\usepackage{tikz}
\usepackage{pgfplots}
\usepackage{lipsum,multicol}

\usepackage{algorithmicx}
\usepackage{algorithm}
\usepackage[noend]{algpseudocode}
\usepackage{setspace}
\newcommand*\Let[2]{\State #1 $\gets$ #2}
\algrenewcommand\algorithmicrequire{\textbf{Input:}}
\algrenewcommand\algorithmicensure{\textbf{Output:}}
\algnewcommand\algorithmicinput{\textbf{Input:}}
\algnewcommand\INPUT{\item[\algorithmicinput]}

\usepackage{subcaption}
\usepackage[pagebackref=true,breaklinks=true,letterpaper=true,colorlinks,bookmarks=false]{hyperref}

\usepackage{xparse}

\NewDocumentCommand\EdgeSet{}{ \mathcal{E} }
\NewDocumentCommand\Rmeas{}{\tilde{\mathbf{R}}_{i,j}}

\newcommand{\frob}[1]{\|#1\|_{\scriptscriptstyle F}}
\newcommand{\mbf}[1]{\mathbf{#1}}

\NewDocumentCommand\bbm{}{ \begin{bmatrix} }
\NewDocumentCommand\ebm{}{ \end{bmatrix} }

\NewDocumentCommand\Transpose{m}{ \left.{#1}\right.^T }

\include{matt_giamou_notation}
\newcommand{\todo}[1]{}

\iccvfinalcopy 

\def\iccvPaperID{2296} 

\ificcvfinal\fi
\begin{document}

\title{Sparse Bounded Degree Sum of Squares Optimization for \\ Certifiably  Globally Optimal Rotation Averaging}


\author{Matthew Giamou\thanks{The corresponding author can be reached at \texttt{matthew.giamou@robotics.utias.utoronto.ca}.},~~ Filip Maric, ~Valentin Peretroukhin, ~Jonathan Kelly\\
University of Toronto
}

\maketitle

\begin{abstract}
Estimating unknown rotations from noisy measurements is an important step in SfM and other 3D vision tasks. Typically, local optimization methods susceptible to returning suboptimal local minima are used to solve the rotation averaging problem. A new wave of approaches that leverage convex relaxations have provided the first certification methods and guarantees for global optimality for state estimation techniques involving SO(3). In this paper, we cast rotation averaging as a polynomial optimization problem over unit quaternions to produce a rotation averaging algorithm that returned certifiably globally optimal solutions for all problem instances tested. This is achieved by formulating and solving a sparse convex sum of squares (SOS) relaxation of the problem. We provide an open source implementation of our algorithm and experiments, demonstrating the benefits of our globally optimal approach.
\end{abstract}

\section{Introduction}
\label{sec:intro}

\begin{figure}[h!]
    \centering
    \includegraphics[width=0.48\textwidth]{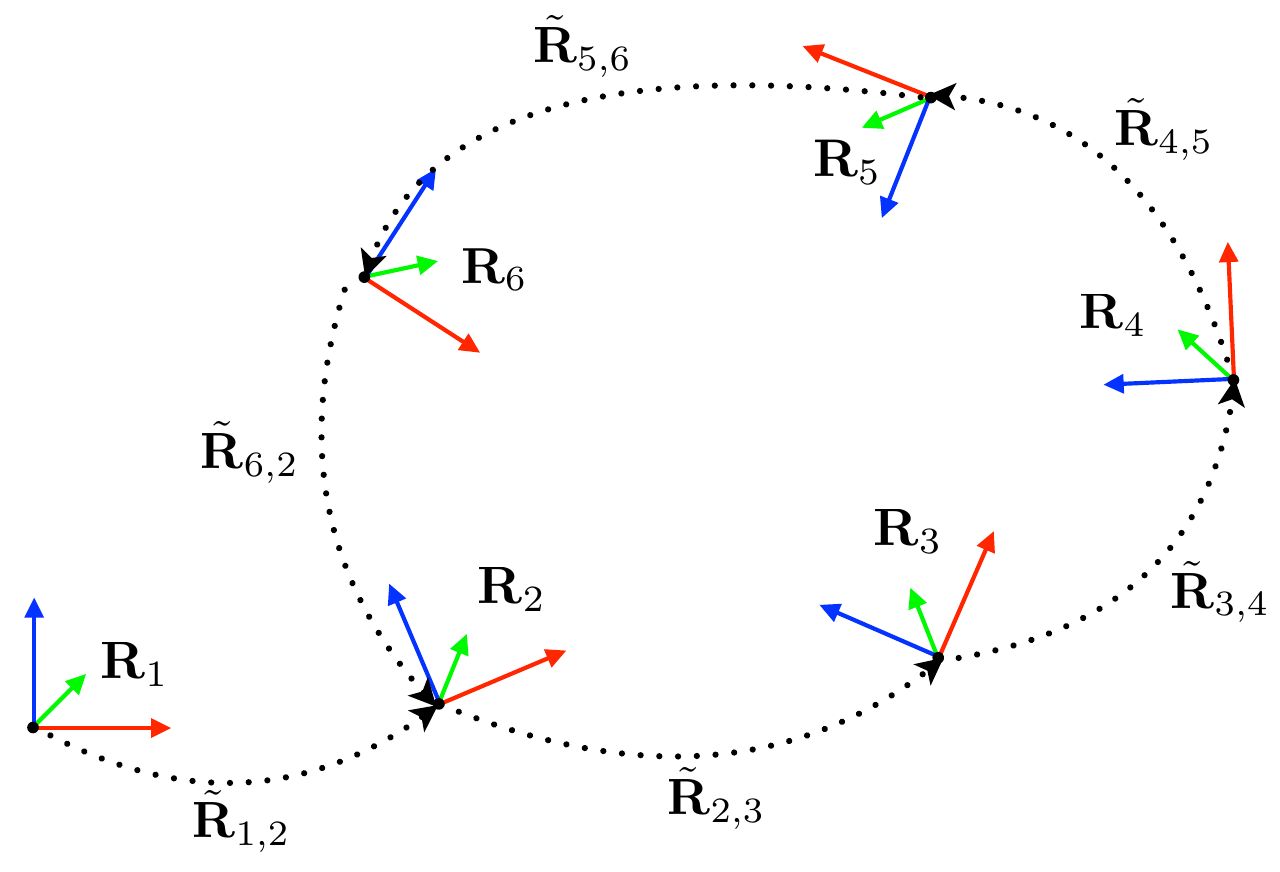}
    \caption{Multiple rotation averaging involves estimating absolute orientations $\mbf{R}_i \in \text{SO(3)}$ given noisy pairwise relative measurements $\tilde{\mbf{R}}_{i,j} \in \text{SO(3)}$ recovered by cameras or other sensors.}
    \label{fig:rotation_averaging}
\end{figure}

Estimating orientations is the principal challenge for many important 3D vision systems. For example, in order to autonomously perform useful tasks, mobile robots need to safely navigate their environment. When a map is unavailable, structure from motion (SfM) or simultaneous localization and mapping (SLAM) is used to build a map while keeping track of a robot's trajectory. The nonlinear and nonconvex constraints of SE(3) and SO(3) variables make solving 3D state estimation problems extremely difficult, and a variety of approaches have been proposed over the past few decades. A combination of probabilistic graphical modelling and local optimization methods has steadily become the \textit{de facto} gold standard approach to solving large-scale 3D SLAM problems \cite{cadena2016past}, but these methods often fail without a good initialization or in the presence of severe noise. However, recent advances in optimization theory have led to \textit{globally optimal} methods that use convex relaxations of state estimation problems \cite{rosen2016se, briales2017cartan, mangelson2018guaranteed, eriksson2018rotation, wang2013exact}. This strategy has led to fast solvers that return solutions that can be \textit{certified} as globally minimal by comparing the gap between the original and relaxed problems' cost functions' values at the candidate solution. 

In many camera-based applications, uncoupling position is a practical strategy, or position is either unobservable or known \cite{carlone2015initialization}. This leads to the SO(3) synchronization or ``rotation averaging" problem, which retains the difficult nonconvex constraints that make full 3D pose estimation challenging. The \textit{multiple} rotation averaging problem (henceforth simply referred to as rotation averaging) involves solving for a number of absolute orientations related by noisy relative measurements (see Figure \ref{fig:rotation_averaging}). A thorough survey of the rotation averaging problem is given in \cite{hartley2013rotation}, wherein two related variants of the problem (single and conjugate rotation averaging) are also examined. 

\subsection{Contributions} 
In this work, we examine rotation averaging through the lens of recent sum of squares (SOS) polynomial optimization research. Our approach is inspired by the algorithm in \cite{mangelson2018guaranteed} for 2D SLAM problems, which makes use of the sparse bounded degree SOS (Sparse-BSOS) optimization method presented in \cite{weisser2018sparse}. Our main contributions are:

\begin{enumerate}
\item a discussion and comparison of polynomial optimization formulations of rotation averaging;
\item and an open source implementation of our certifiably globally optimal algorithm and experiments using the Sparse-BSOS Matlab package\footnote{ \scriptsize \url{https://github.com/utiasSTARS/sos-rotation-averaging}}.
\end{enumerate}
Formal certification methods are essential for autonomous vehicles and other safety-critical applications of computer vision. Our hope is that this first application of SOS optimization to a challenging rotation estimation problem will motivate further development of this powerful family of techniques for estimation problems in vision and robotics. 

\subsection{Related Work}
Globally optimal solutions to 3D vision and state estimation problems were considered in \cite{olsson2008solving} via convex Lagrangian duality-based relaxations. More recently, Lagrangian duality has been applied to SLAM \cite{carlone2015lagrangian, rosen2016se, briales2017cartan}, relative camera pose estimation \cite{briales2018certifiably}, sensor calibration \cite{giamou2019certifiably}, and rotation averaging problems \cite{eriksson2018rotation}. These approaches provide fast solution and verification methods for 3D optimization, but only when measurement error is bounded by a problem instance-specific value that is often not straightforward to compute.

Sum of squares (SOS) optimization is a set of approaches that uses theoretical results in real algebraic geometry to solve feasibility and optimization problems with polynomial variables and constraints~\cite{parrilo2003semidefinite, lasserre2001global}. These techniques have been applied to numerous problems in probability, statistics, control theory, and finance for almost two decades but have only recently found limited application to state estimation problems in computer vision and robotics \cite{lasserre2010moments}. For example, SOS optimization is used in \cite{heller2014hand} to find globally optimal solutions to hand-eye and robot-world calibration problems. It was also applied to the \textit{single} and \textit{conjugate} rotation averaging problems in \cite{ke2015global}. These problems are noteworthy because they are focused on estimating a small, fixed number of variables representing extrinsic calibration parameters, making them amenable to SOS techniques which typically scale poorly with the number of variables and constraints. Our work is similar to these approaches but uses sparsity properties and a newer SOS relaxation hierarchy on the more challenging multiple rotation averaging problem.

More recently, a certifiably globally optimal solution to 2D SLAM (with and without landmarks) was developed in \cite{mangelson2018guaranteed}. A new SOS relaxation hierarchy called Sparse Bounded Degree Sum of Squares (Sparse-BSOS) \cite{weisser2018sparse} was applied to a polynomial optimization formulation of 2D SLAM. Although the runtime is too slow for real-time applications, the algorithm in \cite{mangelson2018guaranteed} is an important step towards using SOS techniques in practical, large-scale robotic state estimation problems. In this paper, we parallel the approach of \cite{mangelson2018guaranteed} to develop a similar algorithm for 3D rotation averaging. In particular, we leverage properties of unit quaternion-based rotation averaging to formulate a sparse polynomial optimization problem. 

\section{Problem Formulation} \label{sec:problem_formulation}
This section describes notation and the rotation averaging problem formulation  used throughout the paper. Quaternionic and chordal cost functions are briefly compared. Polynomial representations of SO(3) constraints are developed to conform with the structure required by Sparse-BSOS in Section \ref{sec:sos}. 

\subsection{Notation}
Boldface is used for vector and matrix variables with lowercase and uppercase letters respectively (e.g., $\mbf{x}, \mbf{A}$). The identity matrix is written $\mbf{I_{n\times n}}$, and the matrix of zeros of size $m\times n$ is denoted $\mbf{0}_{m\times n}$ (or without subscripts when the dimensions can be easily inferred from context). The $p$-norm of a vector $\mbf{x}$ is denoted $\|x\|_p$, with the Euclidean or 2-norm implied when $p$ is omitted. The Frobenius norm of a matrix is denoted $\frob{\mbf{A}}$. We use $[n]$ as shorthand for the set $\{1, \ldots, n\}$. The ring of polynomials in the $n$ scalar variables $x_i$ of $\mbf{x} \in \mathbb{R}^n$ is denoted $\mathbb{R}[\mbf{x}]$. The ring of polynomials only involving some subset of the variables in $\mbf{x}$ indexed by $I \subseteq [n]$ is denoted $\mathbb{R}[\mbf{x}; I]$. 


We make extensive use of unit quaternions to represent absolute orientations and relative orientation measurements. The vector 
\begin{equation} 
\mbf{q} = [q_w\ q_x\ q_y\ q_z]^T \in \mathbb{R}^4 
\end{equation}
is used to represent the quaternion
\begin{equation}
q = q_w + q_x \mbf{i} + q_y \mbf{j} + q_z \mbf{k} \in \mathbb{H}.
\end{equation}
We write a quaternion representing the i$th$ absolute orientation in an ordered set as $\mbf{q}_i$ and write the relative pose between orientations $i$ and $j$ as $\mbf{q}_{i,j}$ such that 
\begin{equation}
\mbf{q}_j = \mbf{q}_i \circ \mbf{q}_{i,j},
\end{equation}
where $\circ$ represents standard quaternion multiplication. Measurements of relative poses are denoted $\tilde{\mbf{q}}_{i,j}$. Note that unit quaternions constitute a ``double cover" of SO(3) in that $\mbf{q}_i$ and $-\mbf{q}_i$ represent the same rotation. Because of this, the quaternion distance metric describing how close two quaternions $\mbf{q}_i$ and $\mbf{q}_j$ are in the ambient Euclidean space is \cite{hartley2013rotation}:
\begin{equation}
\min(\|\mbf{q}_i - \mbf{q}_j\|, \|\mbf{q}_i + \mbf{q}_j\|).
\end{equation}

\subsection{Rotation Averaging}
\todo{ How do we get to here from, say, Langevins noise on the rotation matrix or some other quaternion based uncertainty? Want to avoid a purely geometric approach and instead connect it to MLE/other Bayesian formulation.}
Rotation averaging refers (among other closely related definitions) to the problem of determining a set of absolute orientations (i.e., SO(2) or SO(3) elements) given a set of noisy relative orientation measurements (also represented by SO(2) or SO(3) elements). An instance of the rotation averaging problem consists of a set of relative orientation measurements $\{\Rmeas\}, (i,j) \in \EdgeSet$ where $\EdgeSet$ is a set of $M$ directed edges in the graph relating absolute orientation variables $\mathbf{R}_i, \ i=1, \ldots, N$. We can derive a maximum likelihood estimate (MLE) if we assume each noisy measurement $\Rmeas$ is generated by a Langevins distribution \cite{carlone2015lagrangian} with concentration parameter $\omega_{i,j}^2$:
\begin{align}
\Rmeas &= \Transpose{\mathbf{R}_i}\mathbf{R}_j \mathbf{R}_\epsilon, \\
\mathbf{R}_\epsilon &\sim \text{Langevins}(\mathbf{I}, \omega_{i,j}^2).
\end{align}
The MLE for this problem can be written as a minimization of the sum of the negative log-likelihoods of each independent measurement:
\begin{equation}\label{eq:problem_mle}
f^\star = \underset{\mathbf{R}_i \in \text{SO}(3)}{\text{minimize}} \sum_{(i,j) \in \EdgeSet} -\log \mathcal{L}(\Rmeas | \mathbf{R}_i, \mathbf{R}_j).
\end{equation}
The log-likelihood of a Langevins distributed error is analogous to that of the Gaussian distribution for linear systems in that it has the following desirable least-squares form:
\begin{equation} \label{eq:cost_mle}
-\log \mathcal{L}(\Rmeas | \mathbf{R}_i, \mathbf{R}_j) = \frac{\omega_{i,j}^2}{2}\frob{\mathbf{R}_i \Rmeas - \mathbf{R}_j}^2,
\end{equation}
which makes this formulation of rotation averaging a quadratically constrained quadratic program (QCQP) when the design variables used are the full rotation matrices $\mathbf{R}_i \in \mathbb{R}^{3\times3}$. The norms in the cost function are also known as \textit{chordal} distances \cite{hartley2013rotation}, and they are related to the angular or \textit{geodesic} distances $\theta = \|\log(\mathbf{R}_a^T\mathbf{R}_b) \|$ as follows:
\begin{equation}
\frob{\mbf{R}_a - \mbf{R}_b} = 2\sqrt{2} \sin(\theta/2).
\end{equation}
Similarly, the quaternion distance is related to $\theta$ via:
\begin{equation}
\min(\|\mbf{q}_a - \mbf{q}_b\|, \|\mbf{q}_a + \mbf{q}_b\|) = 2 \sin(\theta/4).
\end{equation}
As we will demonstrate in Section \ref{sec:sos}, we are interested in polynomial cost functions and constraints, which the chordal or MLE formulation of \ref{eq:problem_mle} has been shown to exhibit (SO(3) constraints can be written as quadratic equations). In order to make the quaternion distance quadratic, we can fix the ``sign" $\epsilon_{i,j} \in \{-1, 1\}$ of all measurement quaternions $\tilde{\mbf{q}}_{i,j}$ to get the following QCQP:

\begin{equation} \label{eq:problem_quaternion}
\begin{aligned}
\underset{\mbf{q}_i}{\text{minimize}} \sum_{(i,j) \in \mathcal{E}} & \| \mbf{q}_i \circ (\epsilon_{i,j}\tilde{\mbf{q}}_{i,j}) - \mbf{q}_j\|^2, \\
\text{s.t.  } \ \ \ \ \  \mbf{q}_i^T\mbf{q}_i &= 1, \ i = 1, \ldots, N.\\
\end{aligned}
\end{equation}
Note that with $M$ measurements, there are $2^M$ distinct cost functions possible. In Section \ref{sec:double_cover} we describe the method we use to make a favourable selection of quaternion signs $\epsilon_{i,j}$.

We can also write the cost function of \ref{eq:problem_quaternion} as
\begin{equation}
f = \sum_{(i,j) \in E} f_{i,j},
\end{equation}
where each term is 
\begin{equation} \label{eq:cost_function}
\begin{aligned}
f_{i,j} &= \|\mbf{Q}_{i,j} \mbf{q}_i - \mbf{q}_j \|^2, \\
&= \|[\mbf{Q}_{i,j} \ {-\mbf{I}}] [\mbf{q}_i^T \ \mbf{q}_j^T]^T \|^2, \\
&= [\mbf{q}_i^T \ \mbf{q}_j^T] \mbf{A} [\mbf{q}_i^T \ \mbf{q}_j^T]^T,
\end{aligned}
\end{equation}
where 
\begin{equation}
\begin{aligned}
\mbf{A} &= [\mbf{Q}_{i,j} \ {-\mbf{I}}]^T [\mbf{Q}_{i,j} \ {-\mbf{I}}], \\
&= \begin{bmatrix}
\mbf{I} & -\mbf{Q}_{i,j}^T \\
-\mbf{Q}_{i,j} & \mbf{I}
\end{bmatrix},
\end{aligned}
\end{equation}
and
\begin{equation}
\mbf{Q}_{i,j} =  
\epsilon_{i,j}\begin{bmatrix}
q_{w,i,j} & -q_{x,i,j} & -q_{y,i,j} & -q_{z,i,j} \\
q_{x,i,j} & q_{w,i,j} & q_{z,i,j} & -q_{y,i,j} \\
q_{y,i,j} & -q_{z,i,j} & q_{w,i,j} & q_{x,i,j} \\
q_{z,i,j} & q_{y,i,j} & -q_{x,i,j} & q_{w,i,j}
\end{bmatrix},
\end{equation}
is the right multiplication matrix composed of elements of $\tilde{\mbf{q}}_{i,j}$ such that $\mbf{Q}_{i,j} \mbf{q}_i = \mbf{q}_i \circ \tilde{\mbf{q}}_{i,j}$. Note that the unit norm equality constraint of a quaternion can be rewritten as the following equivalent inequalities:
\begin{equation} \label{eq:unit_constraint_bound}
\begin{aligned}
0 &\leq 1 - \mbf{q}_i^T \mbf{q}_i \leq 1, \\
0 &\leq 2 - \mbf{q}_i^T \mbf{q}_i \leq 1.
\end{aligned}
\end{equation}
This simple way of writing our equality constraint with 4 inequality constraints (2 of which are redundant) will be used by the Sparse-BSOS algorithm in Section \ref{sec:sos}.  We can now write Problem \ref{eq:problem_quaternion} in the following form:
\begin{equation} \label{eq:problem_SBSOS}
\begin{aligned}
\underset{\mbf{q}_i}{\text{minimize}} \ &f, \\
\text{s.t. } \ \ \ \ \ &0 \leq g_j \leq 1  \  \forall j \in [M],
\end{aligned}
\end{equation}
where the $M =  2N$ constraints capture the unit norm of each unit quaternion variable. Finally, note that there are infinite solutions to Problem \ref{eq:problem_SBSOS} due to the gauge symmetry of the problem: for any solution $\{\mbf{q}_i\}_{i \in [N]}$ and any unit quaternion $\mbf{q}$, we can apply $\mbf{q} \circ \mbf{q}_i$ to all $\mbf{q}_i$ to obtain another solution that is also a minimizer of Problem \ref{eq:problem_SBSOS}. We resolve this issue by adding constraints that `anchor' the first variable as $\mbf{q}_1 = [1\ 0\ 0\ 0]^T$.

\section{Sum of Squares Optimization}
In this section, we briefly review sum of squares (SOS) optimization and present the Sparse-BSOS hierarchy used by our rotation averaging algorithm. 
\subsection{SOS Optimization} \label{sec:sos}

Sum of squares (SOS) optimization is an approach for solving polynomial optimization problems. Techniques in the SOS literature leverage results from real algebraic geometry to provide certificates of feasibility for polynomial inequalities. Various SOS-based techniques have been applied to problems in control theory and many branches of engineering since the early 2000s, but they remain under-utilized in the state estimation literature. One reason for this lack of popularity is slow runtime that does not scale well as the number of optimization variables increases. This is a problem for applications like SfM where the number of variables and constraints grow rapidly with the number of measurements available. In this work, we utilize the recent Sparse-BSOS hierarchy of \cite{weisser2018sparse} because it is designed with sparse problems like rotation averaging in mind. The interested reader can examine \cite{parrilo2003semidefinite} and \cite{lasserre2010moments} for in-depth introductions to SOS optimization. The Sparse-BSOS method of \cite{weisser2018sparse} is a sparse extension of \cite{lasserre2017bounded}, which introduced a SOS hierarchy with finite convergence guarantees (i.e., globally optimal solutions will \textit{always} be obtained when certain conditions in the problem setup are met).  

\subsection{Sparse-BSOS}
For a complete treatment of the Sparse-BSOS hierarchy, please refer to \cite{weisser2018sparse} and \cite{mangelson2018guaranteed}. Briefly, we are interested in solving the optimization problem 

\begin{equation}
t^\star = \underset{t \in \mathbb{R}}{\sup} \{t | f(\mbf{x}) - t \geq 0, \ \forall \mbf{x} \in \mbf{K} \},
\end{equation}
where $\mbf{K} = \{\mbf{x} \in \mathbb{R}^{n} | 0 \leq g_j(\mbf{x}) \leq 1, \ j = 1, \ldots, m \}$ is the semialgebraic set defined by our constraints. This is equivalent to solving Problem \ref{eq:problem_SBSOS}. The key insight of SOS optimization is that this problem (and other polynomial optimization problems) can be solved as a semidefinite program (SDP) with Positivstellensatz results from real algebra \cite{lasserre2010moments, parrilo2003semidefinite}. Many SOS relaxation hierarchies have been developed, but we use the sparse bounded-degree SOS (Sparse-BSOS) hierarchy of \cite{weisser2018sparse} because it leverages the sparsity of rotation averaging. The method enforces $f(\mbf{x}) - t \geq 0$ by introducing the function 
\begin{equation}
\begin{aligned}
h_d(\mbf{x}, \pmb{\lambda}) &= \sum_{\alpha, \beta \in \mathbb{N}^m}^{|\alpha|_1 + |\beta|_1 \leq d} \lambda_{\alpha\beta} h_{d, \alpha\beta}(\mbf{x}),\\
h_{d,\alpha \beta}(\mbf{x}) &\vcentcolon = \prod_{j=1}^m g_j(\mbf{x})^{\alpha_j}(1 - g_j(\mbf{x}))^{\beta_j}, \ \mbf{x} \in \mathbb{R}^{n},
\end{aligned}
\end{equation}
where $\pmb{\lambda}$ contains the coefficients $\lambda_{\alpha\beta} \geq 0$ indexed by $\alpha$ and $\beta$, and the parameter $d$ allows us to restrict the number of monomials used to construct $h_d$. Now we seek to optimize 
\begin{equation}
t^\star = \underset{t, \pmb{\lambda}}{\sup} \{t | f(\mbf{x}) - t - h_d(\mbf{x}, \pmb{\lambda}) \geq 0, \ \forall \mbf{x}, \pmb{\lambda} \geq 0 \},
\end{equation} 
where $h_d(\mbf{x}, \pmb{\lambda}) > 0$ when $\mbf{x} \in \mbf{K}$ (see \cite{lasserre2017bounded} for details). Next, the problem is converted to an SDP by restricting the search to $\Sigma[\mbf{x}]_k$, the set of SOS polynomials of degree at most $2k$, which constitute a subset of nonnegative polynomials:
\begin{equation}\label{eq:sparse-bsos}
q_d^k = \underset{t, \pmb{\lambda}}{\sup} \{t | f(\mbf{x}) - t - h_d(\mbf{x}, \pmb{\lambda}) \in \Sigma[\mbf{x}]_k, \ \forall \mbf{x}, \pmb{\lambda} \geq 0 \}.
\end{equation} 
Each $q_d^k$ describes a level of the BSOS hierarchy indexed by $d$ and $k$ \cite{lasserre2017bounded}. 
Finally, to produce the Sparse-BSOS hierarchy we partition Problem \ref{eq:sparse-bsos} into smaller blocks of variables and relevant constraints. These subsets of variables, which we denote $I_l \subseteq [n], \ \forall l \in [p]$, must satisfy a sparsity property called the running intersection property (RIP) described in Section \ref{sec:rip}. 


\section{Algorithm} \label{sec:algorithm}
In this section, we describe our complete approach to solving rotation averaging. This involves ``pre-conditioning" the quaternion representation of measurements to mitigate the effects of the double cover property, providing a partition of the variables that satisfies the RIP, and finally using the Sparse-BSOS solver on particular problem instances. 

\subsection{Quaternion Double Cover} \label{sec:double_cover}
In Section \ref{sec:problem_formulation}, we noted the inherent problem description ambiguity caused by the double cover property of the quaternion representation of SO(3). Since the Sparse-BSOS method requires a purely polynomial cost function, a fixed sign for each quaternion measurement $\tilde{\mathbf{q}}_{i,j}$ must be selected. To this end, we used the method described in Section  of \cite{hartley2013rotation} which is summarized in Algorithm \ref{alg:double_cover}. Although we are unaware of any formal proof demonstrating that this method always leads to an optimal assignment, the method's runtime is linear in $M$ and $N$ and can be empirically shown to return the optimal assignment for small ($N \leq 10$) problem instances, even when severe angular measurement errors (e.g., exceeding $\pi/2 \text{ rad}$) are present. 
\vfill\null
\columnbreak

\begin{algorithm}
  \caption{Quaternion Sign Selection \cite{hartley2013rotation}}
   \label{alg:double_cover}
   \begin{spacing}{1.1}
  \begin{algorithmic}[1]
    \Require{Edge graph $\mathcal{E}$, quaternion measurements $\tilde{\mbf{q}}_{i,j}$}
    \Ensure{Quaternion signs $\epsilon_{i,j} \in \{-1, 1\}, \ \forall (i,j) \in \mathcal{E}$}
    \Function{QuaternionSigns}{$\mathcal{E}$, $\tilde{\mbf{q}}_{i,j}$} 
    \Let{$\mathcal{T}$}{SpanningTree($\mathcal{E}$)} \Comment{\emph{Form a spanning tree of $\mathcal{E}$}}
    \Let{$\mathbf{q}_1$}{$[1; 0; 0; 0]$} \Comment{\emph{Set root vertex to identity}}
	\For{$(i,j) \in \mathcal{T}$}
		\Let{$\mbf{q}_j$}{$\mbf{q}_i \circ \tilde{\mbf{q}}_{i,j}$}
		\Let{$\epsilon_{i,j}$}{1} 
	\EndFor    
    \For{$(i,j) \in \mathcal{E}\backslash\mathcal{T}$}
    		\Let{$\epsilon_{i,j}$}{$\underset{\epsilon \in \{-1, 1\}}{\text{argmin}} \mbf{q}_i \circ \epsilon\tilde{\mbf{q}}_{i,j}$}
    \EndFor
    \State \Return{$\{\epsilon_{i,j}\}_{(i,j) \in \mathcal{E}}$}
    \EndFunction
  \end{algorithmic}
  \end{spacing}
\end{algorithm}

\subsection{The Running Intersection Property} \label{sec:rip}
The RIP holds if there exists $p \in \mathbb{N}$ and subsets $I_l \subseteq [N]$ and $J_l \subseteq [M]$ for all $l \in [p]$ such that:
\begin{itemize}
\item $f = \sum_{l=1}^p f^l$, for some $f^1, \ldots, f^p$ such that \\ $f^l \in \mathbb{R}[\mathbf{x}; I_l], \ l\in [p]$;
\item $g_j \in \mathbb{R}[\mathbf{x}; I_l]$ for all $j \in J_l$ and $l \in \{1, \ldots, p\}$; 
\item $\bigcup_{l=1}^p I_l = [N]$;
\item $\bigcup_{l=1}^p J_l = [M]$;
\item for all $l \in [p-1]$ there exists $s \leq l$ such that \\ $(I_{l+1} \cap \bigcup_{r=1}^l I_r) \subseteq I_s$. 
\end{itemize}
Note that the simple partitioning $p=1$, $I_1 = \{1, \ldots, N\}$ trivially satisfies the RIP but does not retain the sparsity pattern that makes the Sparse-BSOS solution method faster than non-sparse alternatives. The problem of finding the ``optimal" (i.e., in some sense \textit{sparsest}) partitioning that satisfies the RIP is a challenging problem. Like the approach in \cite{mangelson2018guaranteed}, we use the fast but suboptimal junction tree-construction method of \cite{smail2017junction} to produce partitions that satisfy the RIP. This method requires a sequence $(C_1, \ldots, C_k)$ of $I = \{1, \ldots, N\}$ which covers $I$ (i.e., $\bigcup_{i=	1}^k C_i = I$) as input. In order for the first requirement above (i.e., all variables involved in monomial $f^l$ can be found in the partition $I_l$) to hold, we use the edges $\mathcal{E}$ as our covering sequence. This quick and simple choice may affect the quality of the resulting partition, but we leave characterizing the effect of the variable partition on performance for future work.  Algorithm \ref{alg:junction_tree} summarizes the method used at a high level; please see our open source implementation in Matlab for further details. 

\begin{algorithm}
  \caption{Satisfy RIP via Junction Tree Algorithm \cite{smail2017junction} }
   \label{alg:junction_tree}
   \begin{spacing}{1.1}
  \begin{algorithmic}[1]
    \Require{Variable edge graph $\mathcal{E}$, variable partition $\{I_l\}_{l=1}^p$}
    \Ensure{Variable/constraint partitions $\{I_l, J_l\}_{l=1}^p$ satisfying RIP}
    \Function{SatisfyRIP}{$\mathcal{E}$} 
	\Let{$(C_1, \ldots, C_m)$}{SortByMinVertex($\mathcal{E}$)}
	\Let{$(Q_1, \ldots, Q_{r'})$}{Merge($(C_1, \ldots, C_m)$} \Comment{\emph{Form unions with $C_i$ that share their smallest element}}
	\Let{$(Q_1, \ldots, Q_r)$}{MakeProperSequence($(Q_1, \ldots, Q_{r'})$)} \Comment{\emph{Clean the sequence by removing any subsets of other sets}}
	\Let{$(C_1, \ldots, C_k)$}{JunctionTree($(Q_1, \ldots, Q_r)$}
	\Let{$\{I_l\}_{l=1}^p$}{MakeProperSequence($(C_1, \ldots, C_k)$)}
	\Let{$\{J_l\}_{l=1}^p$}{MapToConstraints($\{I_l\}_{l=1}^p$)}
    \State \Return{$\{I_l\}_{l=1}^p$, $\{J_l\}_{l=1}^p$}
    \EndFunction
  \end{algorithmic}
  \end{spacing}
\end{algorithm}

\subsection{Sparse-BSOS Rotation Averaging}
Our complete Sparse-BSOS Rotation Averaging method is presented in Algorithm \ref{alg:sparse_bsos}. It takes as input a specification of Problem \ref{eq:problem_quaternion} in the form of an edge graph $\mathcal{E}$ and corresponding noisy measurements $\tilde{\mbf{q}}_{i,j}$. The measurement signs $\epsilon_{i,j}$ are selected with Algorithm \ref{alg:double_cover} and applied to $\tilde{\mbf{q}}_{i,j}$. Next, a variable partition $\{I_l\}_{l=1}^p$ covering $[N]$ that satisfies the RIP described in Section \ref{sec:rip} is found with Algorithm \ref{alg:junction_tree}. Finally, the polynomial program in Problem \ref{eq:problem_SBSOS} is instantiated and solved with the Sparse-BSOS software package of \cite{weisser2018sparse}. 

\begin{algorithm}
  \caption{Sparse-BSOS Rotation Averaging}
   \label{alg:sparse_bsos}
   \begin{spacing}{1.1}
  \begin{algorithmic}[1]
    \Require{Variable edge graph $\mathcal{E}$, quaternion measurements $\tilde{\mbf{q}}_{i,j}$}
    \Ensure{Globally optimal solution $\{\mbf{q}_i^\star\}_{i=1}^{N}$}
    \Function{JunctionTree}{$\mathcal{E}$, $\{I_l\}_{l=1}^p$} 
		\Let{$\{\epsilon_{i,j}\}_{(i,j) \in \mathcal{E}}$}{QuaternionSigns($\mathcal{E}$, $\tilde{\mbf{q}}_{i,j}$)}
		\For{$(i,j) \in \mathcal{E}$}
			\Let{$\tilde{\mbf{q}}_{i,j}$}{$\epsilon_{i,j}\tilde{\mbf{q}}_{i,j}$}
		\EndFor
		\Let{$\{I_l\}_{l=1}^p$, $\{J_l\}_{l=1}^p$}{SatisfyRIP($\mathcal{E}$)}
		\Let{$\{\mbf{q}_i^\star\}_{i=1}^{N}$}{Sparse-BSOS($\mathcal{E}$, $\tilde{\mbf{q}}_{i,j}, \{I_l, J_l\}_{l=1}^p$)}
		\State \Return{$\{\mbf{q}_i^\star\}_{i=1}^{N}$}
    \EndFunction
  \end{algorithmic}
  \end{spacing}
\end{algorithm}
\vspace{-0.15in}
\section{Experiments}
In this section, we provide results on experiments using synthetic data (see the supplementary material for more extensive results). We use multiple simulated problem instances with perfect ground truth to demonstrate that our approach returns the global optimum in all cases, and to demonstrate the advantages of our approach. For each random problem instance, $N$ rotation matrices indexed from $1, \ldots, N$ are uniformly sampled and edges connecting connecting rotations $i$ and $i+1$ are added to the edge graph $\mathcal{E}$. Then, $n_l$ `loop closures' between non-consecutive rotations are introduced by randomly sampling pairs without replacement. The notion of a loop closure is used in robotics to indicate when a robot has returned to a location it previously observed. Thus, each randomly generated rotation averaging problem instance has $N-1 + n_l$ edges representing pairwise measurements. 

In all experiments, error is introduced to idealized ground truth rotation matrix measurements via an axis-angle perturbation where the axis is uniformly sampled over the unit sphere, and the magnitude of rotation is drawn from a uniform distribution over $[-\theta_{\text{max}}, \theta_{\text{max}}]$. This noise injection strategy was chosen because unlike the chordal loss of \ref{eq:cost_mle}, the quaternionic loss is not a MLE estimate. Additionally, the uniform distribution, unlike a Gaussian or von Mises distribution, provides a hard bound on the maximum single measurement error present in a problem instance. In all cases presented, semidefinite programs (SDPs) are solved by the SDPT3 Matlab software \cite{toh1999sdpt3}. This is a popular, generic SDP solver that is not specifically tailored to problems exhibiting any specific structure (see Section \ref{sec:runtime} for further discussion). 

\subsection{Global Optimality}
In every single simulated problem instance, our approach (Algorithm \ref{alg:sparse_bsos}) produced a certifiably global minimum. The Lagrangian duality-based approach of Fredriksson et al. in \cite{fredriksson2012simultaneous} also returned the global minimum in all tested instances. Please see the supplementary material for detailed results and further discussion.

\subsection{Error Analysis} \label{sec:error_analysis}
\begin{figure*}[h!]
\centering
\includegraphics[width=\textwidth]{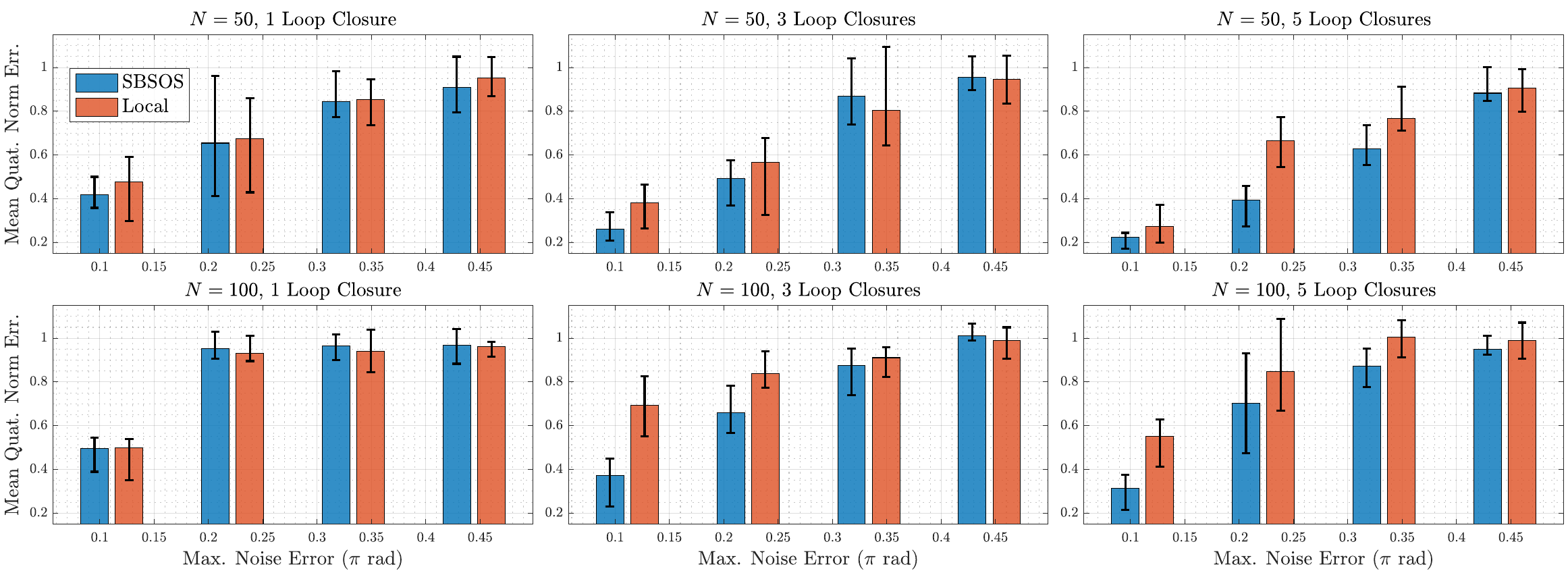}
\caption{Mean quaternion norm error for SBSOS and Local methods. Each bar chart specifies a value of $N$ and $n_l$, while each individual bar represents the mean over 10 random runs for a value of $\theta_{\text{max}}$, which varies along the x-axis. The error bars represent the 1st and 3rd quartile over the 10 random runs.}
\label{fig:error_results}
\vspace{-0.13in}
\end{figure*}

To investigate the accuracy benefits of a globally optimal approach, we use the following mean quaternion norm error between an estimate $\{\mbf{q}_i^\star\}_{i=1}^{N}$ and the ground truth $\{\mbf{q}_i\}_{i=1}^{N}$:
\begin{equation} \label{eq:mean_quaternion_norm_error}
\frac{1}{N}\sum_{i=1}^N \min(\|\mbf{q}_i - \mbf{q}_i^\star\|_2, \|\mbf{q}_i + \mbf{q}_i^\star\|_2).
\end{equation}
Figure \ref{fig:error_results} compares the performance in terms of this metric for our approach (labelled `SBSOS') and a common local optimization method (labelled `Local'). In the local approach, the measurements corresponding to sequential edges $i, i+1$, along with the fixed initial pose vlaue $\mbf{q}_0$ are used to initialize the values of each $\mbf{q}_i$. Then, Matlab's local \texttt{fmincon} function is used to locally optimize the cost function. Our globally optimal approach has markedly better accuracy when the maximum angular error $\theta_{\text{max}} \leq 0.5\pi\text{ rad}$. Additionally, increasing the graph density (via $n_l$) also improves the relative performance of SBSOS over Local. 

\subsection{Runtime} \label{sec:runtime}
In this section, we compare the runtime of our algorithm and the SDP relaxation of Fredriksson et al. found in \cite{fredriksson2012simultaneous}. Figure \ref{fig:runtime_low_noise} compares the mean runtime of each approach over 10 runs for a variety of $N$ and $n_{l}$ values. For the parameter range displayed, it is clear that our approach exploits sparsity to run faster than the standard SDP relaxation (labelled `Fredriksson'), which scales very poorly with $N$. However, for values of $n_l$ greater than 5, the mean and variance of the SBSOS runtime begins to increase rapidly, often taking much longer than the standard SDP. Further investigation into choosing a partition over dense edge graphs is needed to ensure a reasonable runtime. However, large sparse graphs are common in robotics applications where a vehicle navigating a large environment infrequently returns to previously visited areas.
The runtime of local methods like the one described in Section \ref{sec:error_analysis} is much faster (e.g., less than a second for over a hundred poses in Matlab) than our certifiably globally optimal approach, which does not scale well with graph density. However, there is great promise that fast block-coordinate minimization (BCM) algorithms like the one used for chordal rotation averaging in \cite{eriksson2018rotation} can be applied to our approach and Fredriksson et al.'s SDP relaxation. Recent work has provided formal guarantees that these methods can find the global minimum to SDPs formed by relaxing problems involving unit norm and orthogonality constraints in a fraction of the time taken by interior-point methods like SDPT3 \cite{tian2019block, erdogdu2018convergence}. In this paper's sequel, we intend to apply these methods (and their associated theoretical machinery) to both the standard SDP relaxation and the SDP resulting from our application of the Sparse-BSOS hierarchy to rotation averaging. 
\begin{figure}
	\centering
	\resizebox{0.8\columnwidth}{!}{\input{fig/graph_closures_N.tex}}
	\caption{Comparison of solver runtimes for Fredriksson's SDP relaxation \cite{fredriksson2012simultaneous} and SBSOS for $\theta_{\text{max}} = \frac{2\pi}{10}$. Each bar displays the mean runtime in seconds over 10 runs.}
	\label{fig:runtime_low_noise}
	\vspace{-0.2in}
\end{figure}

\section{Conclusion and Future Work}
In this paper, we have presented the first application of sum of squares optimization techniques to 3D rotation averaging, resulting in an algorithm that found a certifiable global minimum of all problem instances tested. 

In addition to further experimentation on real and synthetic data, a thorough investigation into block-coordinate minimization methods for convex relaxations of rotation averaging will lead to faster runtimes \cite{tian2019block, erdogdu2018convergence}. Finally, a comparison with other rotation averaging schemes and cost functions is needed, especially those that have optimality guarantees~\cite{eriksson2018rotation}. 

%

%

{\small
\bibliographystyle{ieee}
\bibliography{sos}
}

\end{document}


\title{Supplementary Material for ``Sparse Bounded Degree Sum of Squares Optimization for Certifiably  Globally Optimal Rotation Averaging"}

\author{Matthew Giamou\thanks{The corresponding author can be reached at \texttt{matthew.giamou@robotics.utias.utoronto.ca}.},~~ Filip Maric, ~Valentin Peretroukhin, ~Jonathan Kelly\\
University of Toronto
}
\maketitle


\section{Introduction}
\label{sec:intro}
This supplementary material contains some figures and comments summarizing experimental results that were unable to fit in the main body of the paper ``Sparse Bounded Degree Sum of Squares Optimization for Certifiably  Globally Optimal Rotation Averaging". Specifically, more data on algorithm runtimes and global optimality are provided.

\section{Runtime}
\begin{figure}
	\centering
	\resizebox{0.8\columnwidth}{!}{\input{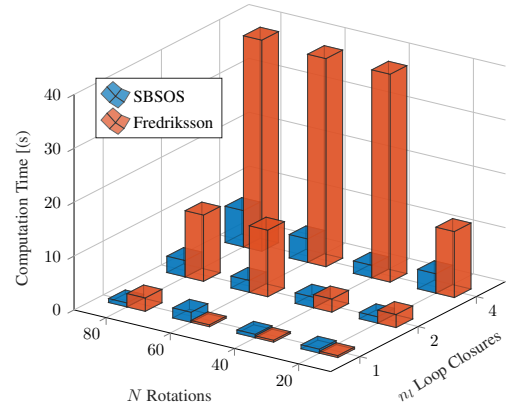}}
	\caption{Comparison of solver runtimes for Fredriksson et al.'s SDP relaxation \cite{fredriksson2012simultaneous} and our SBSOS method for $\theta_{\text{max}} = \frac{9\pi}{10}$. Each bar displays the mean runtime in seconds over 10 runs.}
	\label{fig:runtime_high_noise}
\end{figure}

In addition to the lower noise ($\theta_{\text{max}} = 2\pi/10$) case in Figure 3 of the main paper, we present the noiser $\theta_{\text{max}} = 9\pi/10$ case in Figure \ref{fig:runtime_high_noise}. Once again, SBSOS scales much better with respect to the number of variables $N$ than the standard SDP relaxation of Fredriksson et al., but struggles with increasing graph density (i.e., when the number of loop closures $n_l$ is large). However, the trend of increasing SBSOS runtime for increasing $n_l$ is a bit less severe in the noiser case of Figure \ref{fig:runtime_high_noise}. We do not have an explanation for this behaviour at the moment but intend to investigate this phenomenon in future work. 

\section{Global Optimality}
\begin{figure*}[h!]
\centering
\includegraphics[width=\textwidth]{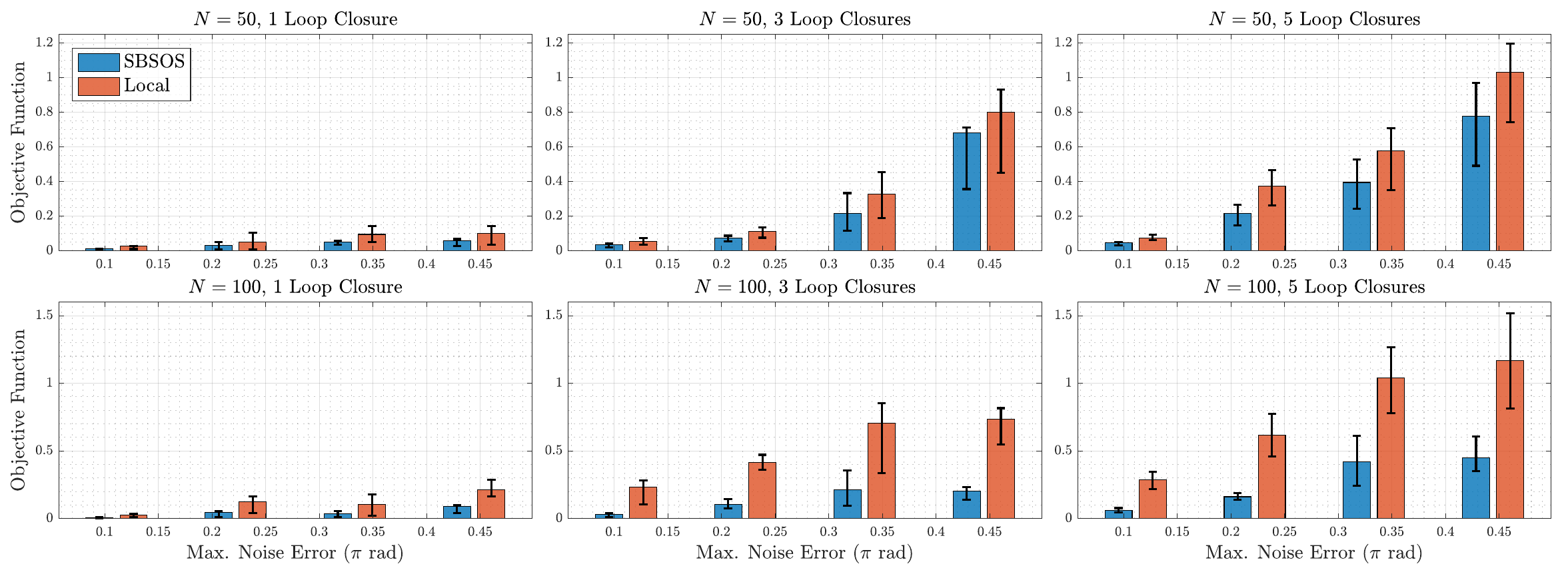}
\caption{Mean objective function values for SBSOS and Local methods. Each bar chart specifies a value of $N$ and $n_l$, while each individual bar represents the mean value of the objective function over 10 random runs for a value of $\theta_{\text{max}}$, which varies along the x-axis. The error bars represent the 1st and 3rd quartile over the 10 random runs. The values returned by SBSOS were certified as globally optimal via the absence of an objective function gap between the original and relaxed problems, whereas the local method is clearly returning solutions corresponding to local minima.}
\label{fig:cost_function_results}
\vspace{-0.13in}
\end{figure*}

In the main paper, we mentioned that the Sparse-BSOS method and SDP relaxation both returned the same certifiably globally optimal minimum for all test cases. In this section, we compare the cost function of the local method with the globally optimal cost function value. 
Figure \ref{fig:cost_function_results} displays the mean objective function values for the same set of random runs displayed in Figure 2 of the main paper. As predicted by the theory dictating that SBSOS will always return the global minimum, the local method is worse in every single case. This verifies the hypothesis that the worse error results at low noise displayed in Figure 2 of the main paper are due to the local method becoming trapped in local minima. At higher noise levels, the globally optimal method still finds much better solutions in terms of the objective function, but this is not strongly reflected in the actual error as the severe noise erodes the value of the true global minimum. 
The experiments in \cite{fredriksson2012simultaneous} provide a use case for global optimality in high-error problem instances in the context of an iterative-reweighted least squares (IRLS) solver for robust cost functions. Robust cost functions mitigate the effects of outliers which contain arbitrarily high measurement error on optimization problems. The IRLS method approximates robust cost functions with a sequence of weighted least squares problems that are easier to solve. We intend to investigate these methods and attempt to characterize cases where global optimality aids in providing an accurate estimate.


{\small
\bibliographystyle{ieee}
\bibliography{sos}
}